\pdfoutput=1

\documentclass[11pt]{article}

\usepackage{acl}

\usepackage{times}
\usepackage{latexsym}

\usepackage[T1]{fontenc}

\usepackage[utf8]{inputenc}

\usepackage{microtype}

\usepackage{algpseudocode}
\usepackage{algorithm}
\usepackage{caption}
\usepackage{subcaption}
\usepackage{graphicx}
\usepackage{booktabs}

\title{TaskMix: Data Augmentation for Meta-Learning of Spoken Intent Understanding}

\author{Surya Kant Sahu \\
  Skit.ai \\
  The Learning Machines \\
  \texttt{surya.oju@pm.me}}

\begin{document}
\maketitle
\begin{abstract}
Meta-Learning has emerged as a research direction to better transfer knowledge from related tasks to unseen but related tasks. However, Meta-Learning requires many training tasks to learn representations that transfer well to unseen tasks; otherwise, it leads to overfitting, and the performance degenerates to worse than Multi-task Learning. We show that a state-of-the-art data augmentation method worsens this problem of overfitting when the task diversity is low. We propose a simple method, TaskMix, which synthesizes new tasks by linearly interpolating existing tasks. We compare TaskMix against many baselines on an in-house multilingual intent classification dataset of N-Best ASR hypotheses derived from real-life human-machine telephony utterances and two datasets derived from MTOP. We show that TaskMix outperforms baselines, alleviates overfitting when task diversity is low, and does not degrade performance even when it is high.

\end{abstract}

\begin{figure*}[t]
     \centering
     \begin{subfigure}[b]{0.49\textwidth}
         \centering
         \includegraphics[width=1.0\textwidth]{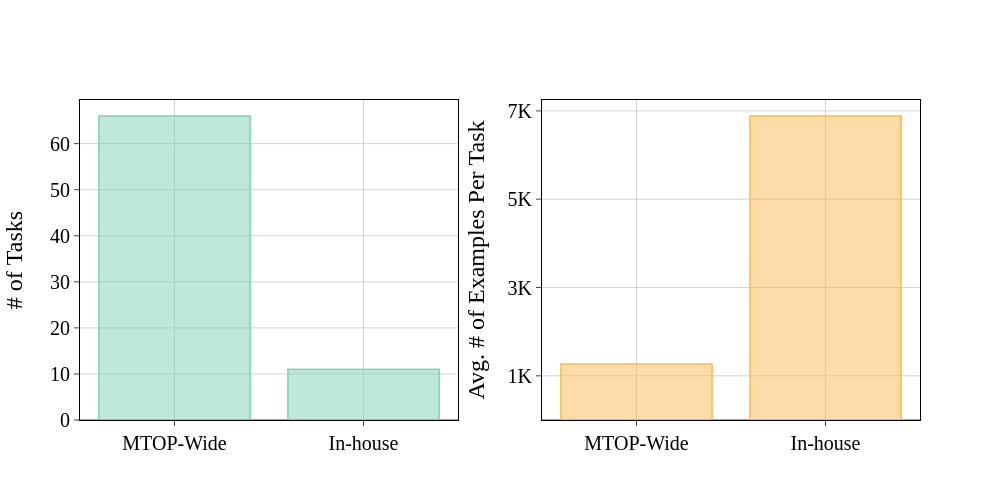}
     \end{subfigure}
     \begin{subfigure}[b]{0.49\textwidth}
         \centering
         \includegraphics[width=\textwidth]{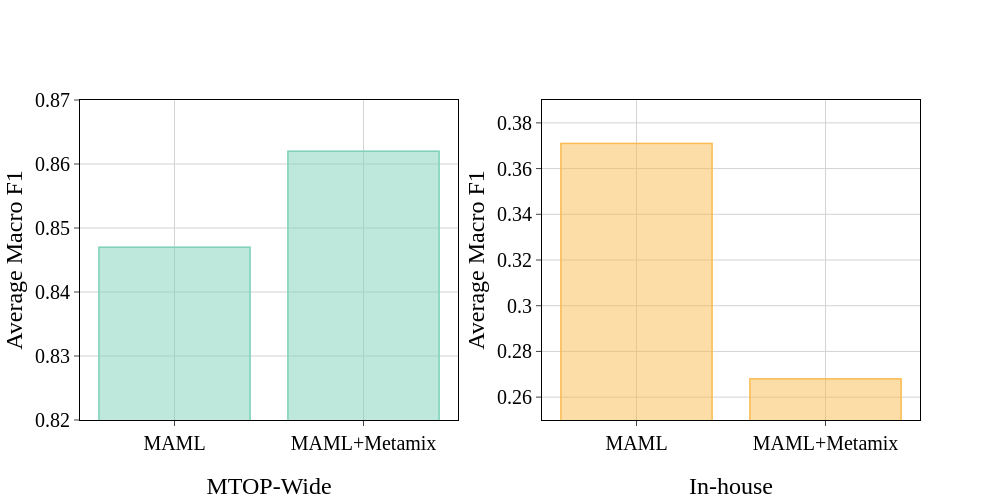}
     \end{subfigure}
    \caption{(Left) Statistics of the two datasets used in this paper. MTOP-Wide has a high \#tasks and a low mean \#examples per task; our In-house dataset has low \#tasks, but a high mean \#examples per task. (Right) Average Macro F1 scores of Model-Agnostic Meta-Learning (MAML) and MAML+MetaMix on both datasets. MetaMix is beneficial for MTOP-Long due to low mean \#examples per task, whereas MetaMix worsens the performance our in-house dataset where the mean \#examples per task is high.}
    
    \label{fig:motivation_plots}
\end{figure*}

\section{Introduction}

Deep learning has seen a meteoric rise in Speech and Language related applications, leading to large-scale applications of Voice-bots, Voice Assistants, Chatbots, etc., which aim to automate mundane tasks such as answering users' queries either in Spoken or textual modality. In many applications, users tend to code-switch or use borrowed words from other languages. A model trained for a particular language will not understand these borrowed words, and hence language-specific models are undesirable in such scenarios. On the other hand, a multilingual model can understand and reason what the user is speaking. 

Due to the scale of the applications, data captured from various sources have different distributions or have different use-cases. Recently, Meta-Learning has emerged as a novel research direction that aims to leverage knowledge from diverse sets of data to learn a transferable initialization so that a low amount of training data is required to adapt to new datasets or tasks.

However, Meta-Learning requires a large number of training tasks, or else the model would overfit to the training tasks and would not generalize well to new tasks \cite{pmlr-v139-yao21b}. In this work, we propose a novel Data Augmentation method, \emph{TaskMix} for meta-learning problems, inspired by MixUp \cite{Zhang2018mixupBE}. We investigate our proposed method against baselines such as MetaMix  \cite{pmlr-v139-yao21b}, Multitask-Learning, and vanilla Transfer Learning for multi-domain multi-lingual Spoken Intent Classification.

\section{Preliminaries}
In this section we describe the problem formulation and the prior work which we built upon.
\subsection{Problem Formulation}

Let $p(\mathcal{T})$ be a distribution over tasks from which training tasks $\mathcal{T}_0, \mathcal{T}_1, \mathcal{T}_2, \dots, \mathcal{T}_{T-1}$ are sampled. The Meta-Learning objective is to learn a model with parameters $\theta$ such that $\theta$ quickly adapts to previously unseen tasks, which are assumed to be sampled from the same underlying distribution $p(\mathcal{T})$; for this paper, each task is a tuple $\mathcal{X}, \mathcal{Y} = \mathcal{T}$, where $\mathcal{X}$ is a set of N-Best hypotheses of utterances, and $\mathcal{Y}$ is a set of corresponding one-hot-encoded intent classes. 

The number of classes in each $\mathcal{Y}$ may differ, and utterances from different $\mathcal{X}$ may be of different language or a different domain. This formulation is general and caters to real-life datasets.

Many meta-learning methods divide each training task into two disjoint sets: support $\mathcal{X}^s, \mathcal{Y}^s$ and query $\mathcal{X}^q, \mathcal{Y}^q$. However, Bai et. al \cite{Bai2021HowII} have shown that a query-set is unnecessary for meta-learning. Hence, throughout this work, we do not split the meta-training tasks, i.e., $\mathcal{X}^s = \mathcal{X}^q = \mathcal{X}$ and $ \mathcal{Y}^s = \mathcal{Y}^q = \mathcal{Y}$

\subsection{Model-Agnostic Meta-Learning}

\begin{algorithm}[t]
\caption{MAML Update, $MetaTrain()$}
\begin{algorithmic}[1]
\Require{$\alpha$: Learning rate for the inner loop.}
\Require{$\beta$: Learning rate for the outer loop.}
\Require{$n$: Iterations for the inner loop.}
\Require{$\mathcal{L}(t, \phi)$: Loss function for task $t$ w.r.t. $\phi$}
    \For {$\mathcal{T}_i \sim p(\mathcal{T})$} \Comment{Sample from support set}
        \State $\theta_i \leftarrow \theta$ \Comment{Copy weights}
        \For {$j = 1$ to $n$}
            \State Evaluate $\nabla_{\theta}\mathcal{L}(\mathcal{T}^s_i, \theta)$
            \State $\theta_{i} \leftarrow \theta_i - \alpha \nabla_{\theta}\mathcal{L}(\mathcal{T}^s_i, \theta)$
        \EndFor
    \EndFor
        \State $\theta \leftarrow \theta - \beta \nabla_{\theta} \sum_{\mathcal{T}^q_i \sim p(\mathcal{T})} \mathcal{L}(\mathcal{T}^q_i, \theta_i)$ \Comment{Update using query set}
\end{algorithmic}
\label{alg:maml}
\end{algorithm}

MAML \cite{Finn2017ModelAgnosticMF} learns the meta-parameters $\theta$ by first, optimizing for multiple steps on a specific task $\mathcal{T}_i$, yielding $\theta_i$ which is the optimal task-specific parameters. This is done for each meta-training task $\mathcal{T}_i \sim p(\mathcal{T})$. Secondly, The loss on the held-out query set is computed, which is back-propagated through the computation graph through each task. Finally, we update $\theta$ such that $\theta$ can be quickly be adapted to each $\theta_i$.

The procedure is outlined in Algorithm \ref{alg:maml}. 

The authors argued that the held-out query set, which isn't used in the inner-loop optimization, prevents the overfitting of task-specific parameters $\theta_i$ and hence improves generalization of meta-parameters $\theta$ to new and unknown tasks.

However, \cite{Bai2021HowII} showed that splitting meta-training tasks into the disjoint query and support sets performs inferior to not splitting at all. Following these results, we do not split and sample data from the same set for inner and outer loops. 




\subsection{MixUp}

MixUp \cite{Zhang2018mixupBE} is a data augmentation technique that synthesizes new datapoints by linearly combining random datapoints in the training set, encouraging a simple, linear behavior between training examples, improving generalization and robustness to noise. The interpolation parameter $\lambda$ is sampled randomly from the Beta distribution at each training step.
As mixing sequences of discrete tokens, such as sentences, is not possible, following \cite{Sun2020MixupTransformerDD}, we only mix the output features of the transformer model.

MetaMix uses MixUp to intra-task datapoints, creating new datapoints within the same task. Whereas our proposed method, TaskMix, extends MixUp to cross-task datapoints, creating \emph{new meta-training tasks}.

\subsection{MetaMix}

\begin{figure*}[t]
     \centering
     \begin{subfigure}[b]{0.28\textwidth}
         \centering
         \includegraphics[width=1.0\textwidth]{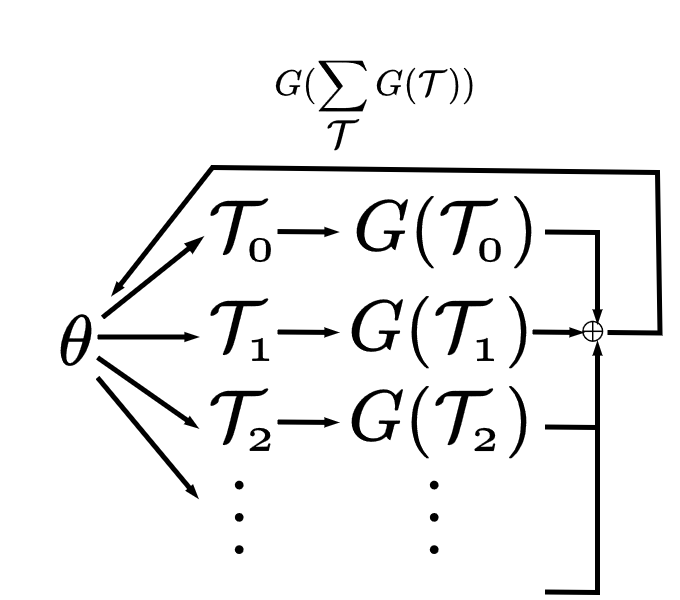}
         \caption{MAML}
     \end{subfigure}
     \hfill
     \begin{subfigure}[b]{0.35\textwidth}
         \centering
         \includegraphics[width=\textwidth]{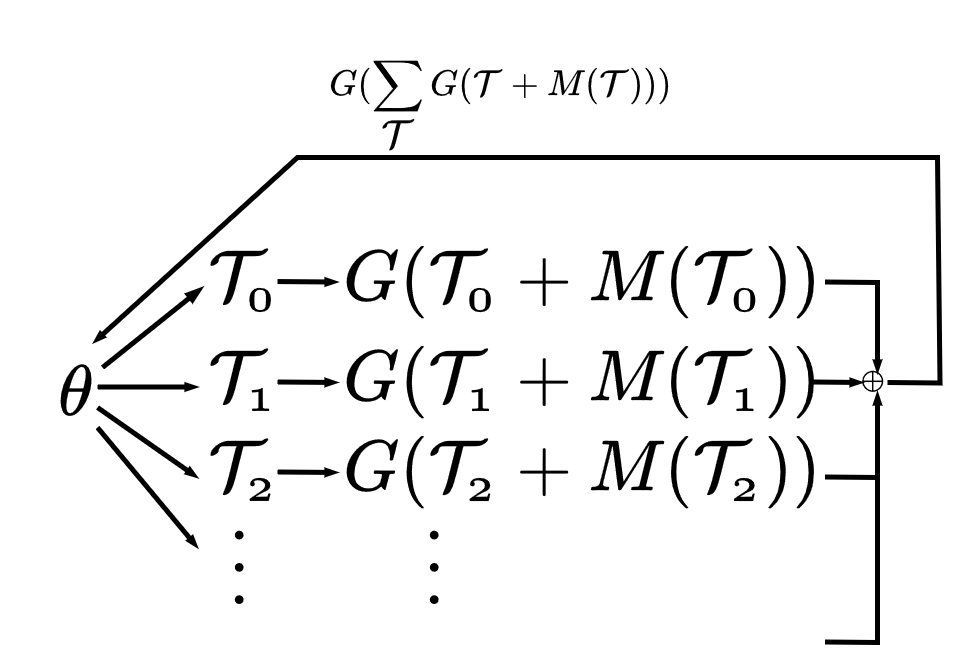}
         \caption{MAML + MetaMix}
     \end{subfigure}
     \hfill
     \begin{subfigure}[b]{0.35\textwidth}
         \centering
         \includegraphics[width=\textwidth]{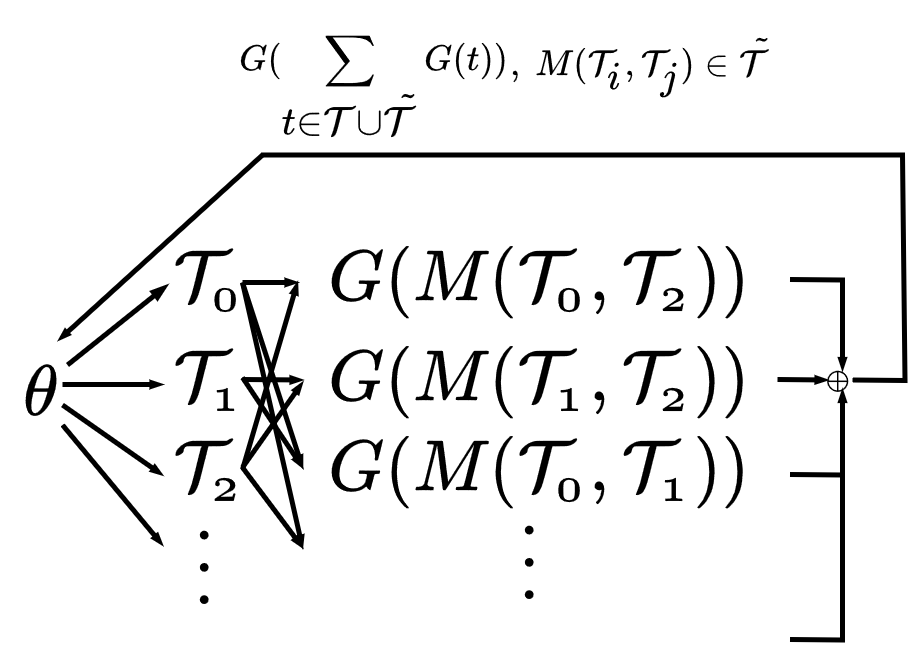}
         \caption{MAML + TaskMix}
     \end{subfigure}
        \caption{Illustration of variants of MAML, including our proposed method, TaskMix. Here, $M(\mathcal{T})$ denotes mixing of datapoints within $\mathcal{T}$; $M(\mathcal{T}_i, \mathcal{T}_j)$ denotes mixing of tasks $\mathcal{T}_i$ and $\mathcal{T}_j$; and $G$ denotes the gradient operator. MetaMix mixes random pairs of datapoints within each task. TaskMix mixes random pairs of tasks at each iteration.}
        \label{fig:maml_variants}
\end{figure*}

MetaMix \cite{pmlr-v139-yao21b} is an application of MixUp to the meta-learning setting. MetaMix encourages generalization within tasks by combining query-set datapoints. Fig. \ref{fig:maml_variants} illustrates how \emph{MAML+MetaMix} differs from \emph{MAML}. MetaMix introduces an additional gradient for each task by mixing random datapoints within each task. MetaMix is a data augmentation method for Meta-Learning where MixUp is applied to random pairs of datapoints \emph{within a batch of query set datapoints} of each task. 

\section{TaskMix}

\subsection{Motivation}
By virtue of meta-learning, $\theta$ learns features that transfer well across tasks \cite{Raghu2020RapidLO}, which requires a large number of meta-training tasks, and most datasets on which studies on Meta-Learning literature use datasets which have a very high number of tasks and low average number of training examples per task. We make the following observations:
\begin{itemize}
    \item MetaMix increases the effective number of datapoints \emph{within each task}, i.e., increasing the mean \#examples per task.
    \item MetaMix does not change the effective number of tasks.
    \item From Fig. \ref{fig:motivation_plots}, we infer that in MTOP-Wide dataset, where the mean \#examples per task is low, MetaMix is very beneficial; however, in our In-house dataset, MetaMix deteriorates performance as the mean \#examples per task is already very high.
    \item Similar to many real-life multi-domain settings, our In-house dataset has a small number of tasks.
\end{itemize}

To this end, we propose a simple data-augmentation method, \emph{TaskMix}, to increase the effective number of tasks used in meta-learning.

\subsection{Method}

We propose a simple method, \emph{TaskMix}, to overcome the low task-diversity problem. First, we sample support and query set batches from all tasks; we then sample $N$ pairs of task indices $\mathbf{I}, \mathbf{J}$ uniformly in the range $[0, T-1]$. For each selected task pair, we sample the interpolation parameter $\lambda$ from the Beta distribution with parameters $(\eta, \eta)$; and then mix the training examples from the support and query sets, resulting in a new synthetic task $\tilde{\mathcal{T}}_n$. Finally, we train with vanilla MAML; however, we train on the new task set $\mathcal{T} \cup \tilde{\mathcal{T}}$. Algorithm \ref{alg:taskmix} describes this procedure.

\begin{algorithm}[t]
\caption{TaskMix}
\begin{algorithmic}[1]
    \Require{$\eta$ : Beta distribution parameter}
    \Require{$mix(a, b, \lambda) = \lambda a + (1 - \lambda) b$}
    \Require{$N$ : Number of new tasks to generate.}
    \While {not converged}
        \For {$t = 0$ to $T-1$}
            \State $x^q_{t} \sim \mathcal{X}^q_{t}, y^q_{t} \sim \mathcal{Y}^q_{t}$ 
            \State $x^s_{t} \sim \mathcal{X}^s_{t}, y^s_{t} \sim \mathcal{Y}^s_{t}$ 
        \EndFor
    
        \State $\mathbf{I}, \mathbf{J} \sim \mathbf{U}^N(0, T-1)$ 
        \For {$i \in \mathbf{I}, j \in \mathbf{J}$, $n = 0$ to $N-1$} 
            \State  $\lambda \sim Beta(\eta, \eta)$ 
            \State $\tilde{x}^q_{n} = mix(x^q_{i}, x^q_{j}, \lambda)$ 
            \State $\tilde{y}^q_{n} = mix(y^q_{i}, y^q_{j}, \lambda)$
            \State $\tilde{x}^s_{n} = mix(x^s_{i}, x^s_{j}, \lambda)$
            \State $\tilde{y}^s_{n} = mix(y^s_{i}, y^s_{j}, \lambda)$
        \EndFor
        \State $MetaTrain()$ 
    \EndWhile
\end{algorithmic}
\label{ticket-search}
\label{alg:taskmix}
\end{algorithm}

TaskMix interpolates between batches of datapoints of random meta-training tasks. In essence, TaskMix encourages generalization across tasks by synthesizing new tasks, while MetaMix encourages generalization within each task by synthesizing new datapoints within the task. We emphasize that TaskMix increases the effective number of tasks, whereas MetaMix increases the effective number of datapoints within each task. We illustrate this difference in Fig. \ref{fig:maml_variants}. We note that TaskMix and MetaMix are orthogonal, and \emph{both methods can be used at the same time}.

TaskMix introduces only one additional hyperparameter, i.e., the number of synthetic tasks $N$. We found that results are largely insensitive to $N$ if $N > T$, but performance rapidly degrades to the performance of MAML if $N < T$, hence we set $N = T$ for all experiments. We recover MAML if we set $N = 0$.

\section{Experiments}

This section presents empirical results on two multi-lingual and multi-domain datasets. For choice of hyperparamters and other experimental details, please refer to the Appendix.

\subsection{Methods and Baselines} 

We use the N-Best-ASR Transformer \cite{Ganesan2021NBestAT} convention of concatenating N-Best ASR transcription hypotheses and then feed the concatenated text to XLM-RoBERTa \cite{Conneau2020UnsupervisedCR} feature extractor. We use the "base" configuration of pretrained XLM-RoBERTa to extract 768-dimensional vectors of each example for each task. The extracted features are inputs to a \emph{neck}, which is a stack of Linear-Parametric ReLU layers. We chose XLM-RoBERTa as the feature extractor as it is trained on large corpora of multi-lingual text.

We now describe the baselines used in the experiments:
\begin{itemize}
    \item \textbf{Multitask Learning (MTL):} we learn a different \emph{linear head} for each meta-training task, and discard these heads after training, and initialize a new head for each meta-testing task.
    \item \textbf{Vanilla Transfer:} we discard all meta-training tasks and finetune directly to each meta-testing task.
    \item \textbf{MAML:} we append a linear layer with the max number of classes in the respective datasets.
\end{itemize}

\begin{table}[t]
\centering
\begin{tabular}{cccc}
\toprule
\textbf{Dataset} & \textbf{\#Tasks} & \textbf{\begin{tabular}[c]{@{}c@{}}Mean\\ \#Classes\end{tabular}} & \textbf{\begin{tabular}[c]{@{}c@{}}Mean\\ \#Examples\\ Per Task\end{tabular}} \\
\midrule
In-house         & 11              & 7.73                   & 6884                             \\
MTOP-Long        & 11              & 2.82                   & 7615                             \\
MTOP-Wide        & 66              & 2.17                   & 1269 \\                            
\bottomrule
\end{tabular}
\caption{Various statistics of datasets used in this paper. \label{tab:dataset_stats}}
\end{table}

\subsection{Datasets}

We briefly summarize the datasets used in this paper. Various statistics relating to the datasets are in Table \ref{tab:dataset_stats}.

\begin{itemize}
     \item \textbf{In-house} dataset is constructed by collecting and automatically transcribing phone calls from various customer-call centers (varying domains, such as restaurants, airlines, banking, etc.) across 3 countries and with conversations comprising at least 3 languages with users speaking with borrowed words, code-switching, etc. Multiple human annotators manually label the intent for each user turn (consisting of 5-Best ASR hypotheses) in a conversation. The resulting dataset contains about 70K utterances across 11 tasks, grouped into 7 meta-training and 4 meta-testing tasks. We grouped the meta-training and meta-testing tasks chronologically, i.e., the oldest 7 tasks were designated as the meta-training tasks and the rest as meta-testing tasks. We use this setup to have as low an application gap as possible.
    \item \textbf{MTOP-Wide \cite{Li2021MTOPAC}} contains over 100K utterances, (which we treat as 1-best hypotheses) from 6 languages across 11 domains. We divide the MTOP dataset by grouping examples from distinct domains and languages, resulting in 66 subsets. We further group these subsets into 54 meta-training and 14 meta-testing tasks. We only keep examples for which the class frequency is at least 50. We create this dataset to have a high task diversity but low average \#examples per task.
    \item \textbf{MTOP-Long \cite{Li2021MTOPAC}} We divide the MTOP dataset by grouping examples from distinct domains resulting in 11 subsets. We further group these subsets into 7 meta-training and 4 meta-testing tasks. We only keep examples for which the class frequency is at least 20. We create this dataset to have a low task diversity but high average \#examples per task.
\end{itemize}

\subsection{Evaluation}
As all tasks across all datasets are highly imbalanced, we use the Macro F1 score to weigh all classes equally.

All tasks are grouped into meta-training and meta-testing sets; each task is split into "support" and "test" sets. For modeling, we first train on the meta-training tasks, then use the same weights to fine-tune on the meta-testing tasks, and then compute Macro F1 scores for each meta-testing task. We then compute the mean of Macro F1 scores across all meta-testing tasks. We denote this metric as \emph{Average Macro F1}. Finally, we report the mean and standard deviation of Average Macro F1 scores across three independent trials with different seeds.

\begin{table}[t]
\centering
\begin{tabular}{ccc}
\toprule
\textbf{Method}               & \textbf{\begin{tabular}[c]{@{}c@{}}Average\\ Macro F1\end{tabular}} \\
\midrule
MTL                           & $0.320 \pm 0.004$                                                      \\
Vanilla Transfer              & $0.321 \pm 0.007$ 
                                           \\

MAML                          & $0.361 \pm 0.021$                                                      \\
MAML+MetaMix                  & $0.265 \pm 0.006$                                                      \\
\underline{MAML+TaskMix}                  & \underline{0.370} $\pm$ \underline{0.023}                                                      \\
\textbf{MAML+MetaMix+TaskMix} & \textbf{0.441 $\pm$ 0.002}                                             \\
\bottomrule
\end{tabular}
\caption{Results on our In-house dataset. We observe that TaskMix yields a significant performance boost. MAML+MetaMix degrades performance to worse than MAML. \label{tab:inhouse_results}}
\end{table}

\begin{table}[t]
\centering
\centering
\begin{tabular}{cc}
\toprule
\textbf{Method}               & \textbf{\begin{tabular}[c]{@{}c@{}}Average\\ Macro F1\end{tabular}} \\
\midrule
MTL                           & $0.439 \pm 0.022$                                                      \\
Vanilla Transfer              & $0.446 \pm 0.014$                                                      \\
MAML                          & $0.442 \pm 0.002$                                                      \\
MAML+MetaMix                  & \underline{0.450} $\pm$ \underline{0.011}                                                      \\
\textbf{MAML+TaskMix}         & \textbf{0.462} $\pm$ \textbf{0.012}                                    \\
MAML+MetaMix+TaskMix          & $0.421 \pm 0.008$                                                      \\

\bottomrule
\end{tabular}
\caption{Results on the MTOP-Long \cite{Li2021MTOPAC} dataset. MAML+TaskMix out-performs other baselines.   \label{tab:mtop2_results}}
\end{table}

\subsection{Results and Discussion}
\label{sec:discussion}

We make the following key observations from Tables \ref{tab:inhouse_results}, \ref{tab:mtop2_results}, and \ref{tab:mtop_results}: 
\begin{itemize}
    \item TaskMix improves performance on "long" datasets i.e., on In-house and MTOP-Long where the \#meta-training tasks are very low and \# examples per task is high.
    \item For the In-house dataset, MetaMix degrades performance to be comparable to vanilla-transfer, i.e., almost no gain from meta-training tasks. We infer that MetaMix makes the model overfit on meta-training tasks, as the number of examples-per-task is already very high.
    \item In any of the datasets, TaskMix \emph{doesn't degrade} the performance of MAML.
    \item For MTOP-Wide, TaskMix only has a slight performance boost compared to other baselines, suggesting that TaskMix is not useful if the number of tasks is already high.
\end{itemize}

We interestingly find that MAML+MetaMix+TaskMix is the worst performing method for MTOP-Long. However, TaskMix is beneficial when used on its own. We leave studying the interaction between MetaMix and TaskMix for future work.

\section{Conclusion}
In this paper, we propose a novel data-augmentation method, TaskMix, to alleviate the problem of overfitting in Meta-learning datasets when the task diversity is too low. Through experiments on two multilingual, multi-domain intent classification datasets, MetaMix could worsen the overfitting problem when the task diversity is low, whereas TaskMix is beneficial in such cases.

\begin{table}[t]
\centering
\centering
\begin{tabular}{cc}
\toprule
\textbf{Method}               & \textbf{\begin{tabular}[c]{@{}c@{}}Average\\ Macro F1\end{tabular}} \\
\midrule
MTL                           & $0.826 \pm 0.018$                                                      \\
Vanilla Transfer              & $0.804 \pm 0.003$                                                      \\
MAML                          & $0.847 \pm 0.006$                                                      \\
\textbf{MAML+MetaMix}                  & \textbf{0.862 $\pm$ 0.006}                                                      \\
\underline{MAML+TaskMix}                  & \underline{0.856} $\pm$ \underline{0.003}                                                      \\
\textbf{MAML+MetaMix+TaskMix} & \textbf{0.861 $\pm$ 0.017}                                             \\

\bottomrule
\end{tabular}
\caption{Results on the MTOP-Wide \cite{Li2021MTOPAC} dataset. MetaMix is beneficial and TaskMix doesn't negatively affect performance (compared to MAML) even when task diversity is high.\label{tab:mtop_results}}
\end{table}

\section{Acknowledgement}
I thank my colleagues at Skit.ai who curated the in-house dataset, brain-stormed, and provided critical and insightful reviews of this work.


\bibliography{anthology,custom}
\bibliographystyle{acl_natbib}

\end{document}


\maketitle
\begin{figure*}[t]
    \centering
    \includegraphics[width=1.0\textwidth]{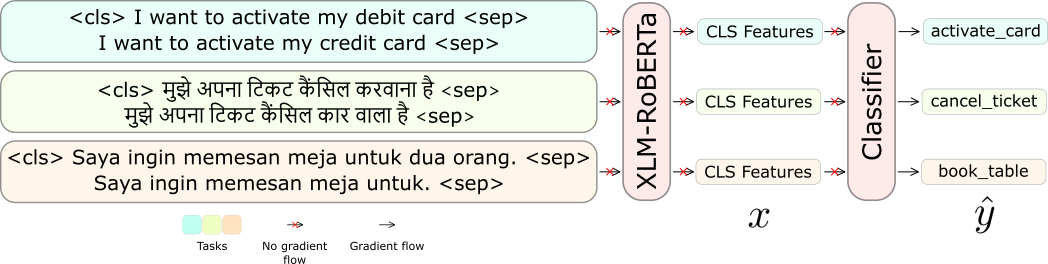}
    \caption{Flow diagram of feature extraction for multi-lingual multi-domain intent classification. For the purposes of this paper, only the \textbf{Classifier} block is optimized; although, the parameters of XLM-RoBERTa can be optimized as well.}
    
    \label{fig:feature_extraction}
\end{figure*}
\appendix

\section{Experimental Details}

The feature extraction process from XLM-RoBERTa is illustrated in Fig. \ref{fig:feature_extraction}.

We implement all baselines and experiments on PyTorch. The experiments were performed on a Ubuntu-based machine with two 24 GB NVIDIA A30 GPUs with CUDA 11.6. We tune hyperparameters for all methods using Optuna. All experiments, except hyperparameter tuning took about one day to run.

We use Adam optimizer for the query-set (outer loop) and SGD for the support set (inner loop). The number of inner-loop steps is $5$ for all MAML-based methods. For TaskMix and MetaMix, we set $\eta = 0.5$ for the Beta distribution parameter. For TaskMix, we set $N = T$, i.e., the number of new synthetic tasks per step equals the number of meta-training tasks for both datasets. We set the batch size to $1024$ for all baselines and use Cosine annealing learning rate with the maximum step as $5000$.

For MTOP dataset, the neck is 3 layers deep and 768 units wide, and for the inhouse dataset, the neck is 24 layers deep and 128 units wide.

As both the datasets are skewed, we use Weighted Cross-Entropy as the loss function for training models, with the weight for each class defined as the inverse class frequency. For TaskMix, we pad each class-weight vector with zeros to the maximum number of classes and mix the class-weight vectors.

For all methods, we early stop the meta-training stage with average task loss of the validation splits of the meta-training tasks; and Average Macro F1 score on the validation splits of the meta-testing tasks, and finally test on the testing splits.

\bibliography{anthology,custom}
\bibliographystyle{acl_natbib}